\documentclass[pdflatex,sn-mathphys-num,oneside]{sn-jnl}

\usepackage{graphicx}%
\usepackage{multirow}%
\usepackage{amsmath,amssymb,amsfonts}%
\usepackage{amsthm}%
\usepackage{mathrsfs}%
\usepackage[title]{appendix}%
\usepackage{xcolor}%
\usepackage{textcomp}%
\usepackage{manyfoot}%
\usepackage{booktabs}%
\usepackage{algorithm}%
\usepackage{algorithmicx}%
\usepackage{algpseudocode}%
\usepackage{listings}%
\usepackage{pdfpages}
\usepackage{amsmath}
\usepackage{amssymb}
\usepackage{makecell}

\theoremstyle{thmstyleone}%
\theoremstyle{thmstyletwo}%

\theoremstyle{thmstylethree}%

\begin{document}

\title[Article Title]{MEDNA-DFM: A Dual-View FiLM-MoE Model for Explainable DNA Methylation Prediction}


\author[1,4]{\fnm{Yi} \sur{He}}
\equalcont{These authors contributed equally to this work.}
\author[2]{\fnm{Yina} \sur{Cao}}
\equalcont{These authors contributed equally to this work.}
\author[3,4]{\fnm{Jixiu} \sur{Zhai}}
\author[1,4]{\fnm{Di} \sur{Wang}}
\author[4]{\fnm{Junxiao} \sur{Kong}}

\author*[4,5]{\fnm{Tianchi} \sur{Lu}}\email{tianchilu4-c@my.cityu.edu.hk}

\affil[1]{\orgdiv{Cuiying Honors College}, \orgname{Lanzhou University}, \orgaddress{\city{Lanzhou}, \state{Gansu}, \country{China}}}
\affil[2]{\orgdiv{School of Management}, \orgname{Lanzhou University}, \orgaddress{\city{Lanzhou}, \state{Gansu}, \country{China}}}
\affil[3]{\orgname{Shanghai Innovation Institute}, \orgaddress{\city{Shanghai}, \country{China}}}
\affil[4]{\orgdiv{School of Mathematics and Statistics}, \orgname{Lanzhou University}, \orgaddress{\city{Lanzhou}, \state{Gansu}, \country{China}}}
\affil[5]{\orgdiv{Department of Computer Science}, \orgname{City University of Hong Kong}, \orgaddress{\city{Kowloon}, \state{Hong Kong}, \country{China}}}

\abstract{
Accurate computational identification of DNA methylation is essential for understanding epigenetic regulation. Although deep learning excels in this binary classification task, its "black-box" nature impedes biological insight. We address this by introducing a high-performance model MEDNA-DFM, alongside mechanism-inspired signal purification algorithms. Our investigation demonstrates that MEDNA-DFM effectively captures conserved methylation patterns, achieving robust distinction across diverse species. Validation on external independent datasets confirms that the model's generalization is driven by conserved intrinsic motifs (e.g., GC content) rather than phylogenetic proximity. Furthermore, applying our developed algorithms extracted motifs with significantly higher reliability than prior studies. Finally, empirical evidence from a Drosophila 6mA case study prompted us to propose a "sequence-structure synergy" hypothesis, suggesting that the GAGG core motif and an upstream A-tract element function cooperatively. We further validated this hypothesis via in silico mutagenesis, confirming that the ablation of either or both elements significantly degrades the model's recognition capabilities. This work provides a powerful tool for methylation prediction and demonstrates how explainable deep learning can drive both methodological innovation and the generation of biological hypotheses.
}

\newpage

\maketitle

\newpage

\section{Introduction}\label{sec1}

DNA methylation stands as a cornerstone of epigenetic regulation, playing an indispensable role in orchestrating gene expression, genomic stability, and cellular differentiation~\cite{smith2013dna, robertson2005dna}. Among diverse chemical modifications, 5-methylcytosine (5mC), N6-methyladenine (6mA), and 4-methylcytosine (4mC) constitute the primary components of this network through distinct biochemical mechanisms~\cite{xiao2018n6, breiling2015epigenetic, lv2021advances}. Specifically, 5mC generally functions as a transcriptional repressor involved in critical processes like genomic imprinting and chromatin organization~\cite{smith2013dna}, whereas 6mA and 4mC are pivotal for DNA replication fidelity, repair, and transposon silencing~\cite{xiao2018n6, oliveira2014interplay}. The dynamic equilibrium of these sites maintains biological homeostasis, while aberrant patterns drive tumorigenesis and pathological states, serving as critical biomarkers for early diagnosis~\cite{robertson2005dna, mikeska2014dna, koch2018analysis}. Consequently, identifying these modifications and their sequence patterns is paramount for decoding complex epigenetic regulatory mechanisms.

While experimental techniques like Whole-Genome Bisulfite Sequencing (WGBS) and SMRT sequencing offer high-resolution landscapes, their prohibitive costs and laborious protocols restrict scalability, necessitating a shift toward computational prediction~\cite{yong2016profiling}. Early approaches predominantly relied on conventional machine learning algorithms, such as SVMs and Random Forests, utilizing handcrafted physicochemical and nucleotide frequency features~\cite{xu18review}. Recently, deep learning integration has catalyzed breakthrough advancements, particularly following the introduction of multi-type methylation prediction by iDNA-MS (2020)~\cite{lv2020idna}. The field rapidly transitioned toward end-to-end frameworks for automatic high-dimensional feature extraction: iDNA-ABF (2022) introduced multi-scale biological language modeling for contextual semantics~\cite{jin2022idna}; Methyl-GP (2025) leveraged pre-trained models for cross-species prediction~\cite{xie2025methyl}; and AutoFE-Pointer (2025) optimized precision via auto-weighted feature extraction~\cite{feng2025autofe}. This evolution marks a significant leap from manual feature engineering to autonomous, semantic-aware representation learning.

However, despite Transformer- and BERT-based architectures continually setting performance benchmarks, their massive parameterization renders them inherently opaque "black boxes," severely impeding the extraction of biological insights. Although recent studies have begun to incorporate interpretability analyses, these efforts remain largely fragmented and exhibit significant limitations. Specifically, existing interpretability endeavors, such as those in Methyl-GP~\cite{xie2025methyl} and iDNA-ABF~\cite{jin2022idna}, are predominantly confined to visualizing attention mechanisms to localize potential regions of interest and sequence motifs. However, such superficial observations relying solely on attention weights are prone to generating false-positive identifications of functional regions~\cite{jain2019attention}. While AutoFE-Pointer ~\cite{feng2025autofe} implemented dynamic feature weighting via pointer networks, the observed fluctuation in interpretability when processing specific methylation types (e.g., 5hmC) exposes the inadequacy of lightweight feature selection strategies in capturing complex long-range dependencies. More critically, these studies suffer from a fundamental methodological deficiency: a lack of deep dissection regarding the model's decision-making mechanism. Elucidating how the model utilizes biological information must be a prerequisite for interpretability, rather than an optional supplement. This omission of the critical attribution link obscures the connection between data features and biological significance, thereby creating a bottleneck in translating computational predictive power into interpretable biological knowledge.

To address these limitations, we present a comprehensive study that bridges high-precision prediction with systematic, mechanism-driven explainability. We first developed MEDNA-DFM, a deep learning model founded on a Dual-View architecture and Feature-wise Linear Modulation (FiLM)~\cite{perez2018film}. Specifically, it synergizes DNABERT-6mer~\cite{ji2021dnabert} and DNABERT-2~\cite{zhou2023dnabert} by utilizing global semantic vectors from the latter to dynamically modulate local fine-grained representations of the former via a FiLM ~\cite{perez2018film} guide, subsequently integrating fused features through a Mixture of Experts (MoE) module~\cite{jacobs1991adaptive, shazeer2017outrageously}. This architecture enables superior performance across diverse DNA methylation types, as demonstrated on 17 benchmark datasets covering 5hmC, 4mC, and 6mA (detailed statistics in Supplementary Table 1)~\cite{lv2020idna}. To transcend the limitations of single-species training, we implemented a Multi-species Joint Fine-tuning strategy. Inspired by the biological consensus that identical methylation types share conserved sequence motifs across species, this approach integrates diverse genomic data to compel the model to capture universal, intrinsic biochemical signatures rather than overfitting to species-specific biases ~\cite{zemach2010genome, fu2015n6, xie2025methyl}. More critically, we developed two novel interpretability algorithms—Contrastive Weighted Gradient Attribution (CWGA) and Contrastive Attention Cohen's d (CAD)—that were directly inspired by our mechanistic analysis of the model's internal decision-making (Supplementary Note 1). By applying these custom-designed algorithms to MEDNA-DFM, we effectively overcome the artifacts and false-positive signals inherent in existing interpretability approaches.

We first validated the superior predictive performance of MEDNA-DFM across 17 benchmarks, confirming that the model achieves robust generalization within specific methylation categories. We further applying the model to external independent datasets, validating that predictions are driven by conserved motifs (e.g., GC content) rather than phylogenetic proximity. Furthermore, by applying our CAD and CWGA algorithms to MEDNA-DFM for Signal Disentanglement and Purification, we distilled high-confidence motifs with statistical significance orders of magnitude higher than prior methods. Crucially, through an in-depth investigation of 6mA regulation in Drosophila, we proposed a new “sequence-structure synergy” hypothesis, positing that the GAGG core sequence ~\cite{zhang2015n6} and an upstream A-tract structural element ~\cite{rohs2010origins} act in a binary cooperative manner to facilitate 6mA modification, thereby offering a coherent explanation for key contradictions in the field ~\cite{o2019sources}. To rigorously test this hypothesis within our computational framework, we designed an in silico mutagenesis strategy.  The results compellingly demonstrate that the simulated ablation of either the GAGG motif or the A-tract significantly impairs the model's predictive performance, while a concurrent mutation causes severe performance decay. This in silico validation not only provides strong computational evidence for their cooperative necessity, but also effectively verifies the faithfulness of our interpretability algorithms in capturing true biological dependencies. In summary, this study provides a robust tool for DNA methylation prediction, MEDNA-DFM, as well as a systematic computational framework to generate and validate mechanism-driven biological insights.

\section{Results}\label{sec2}
\begin{figure}[p]
    \centering
    \includegraphics[width=\textwidth]{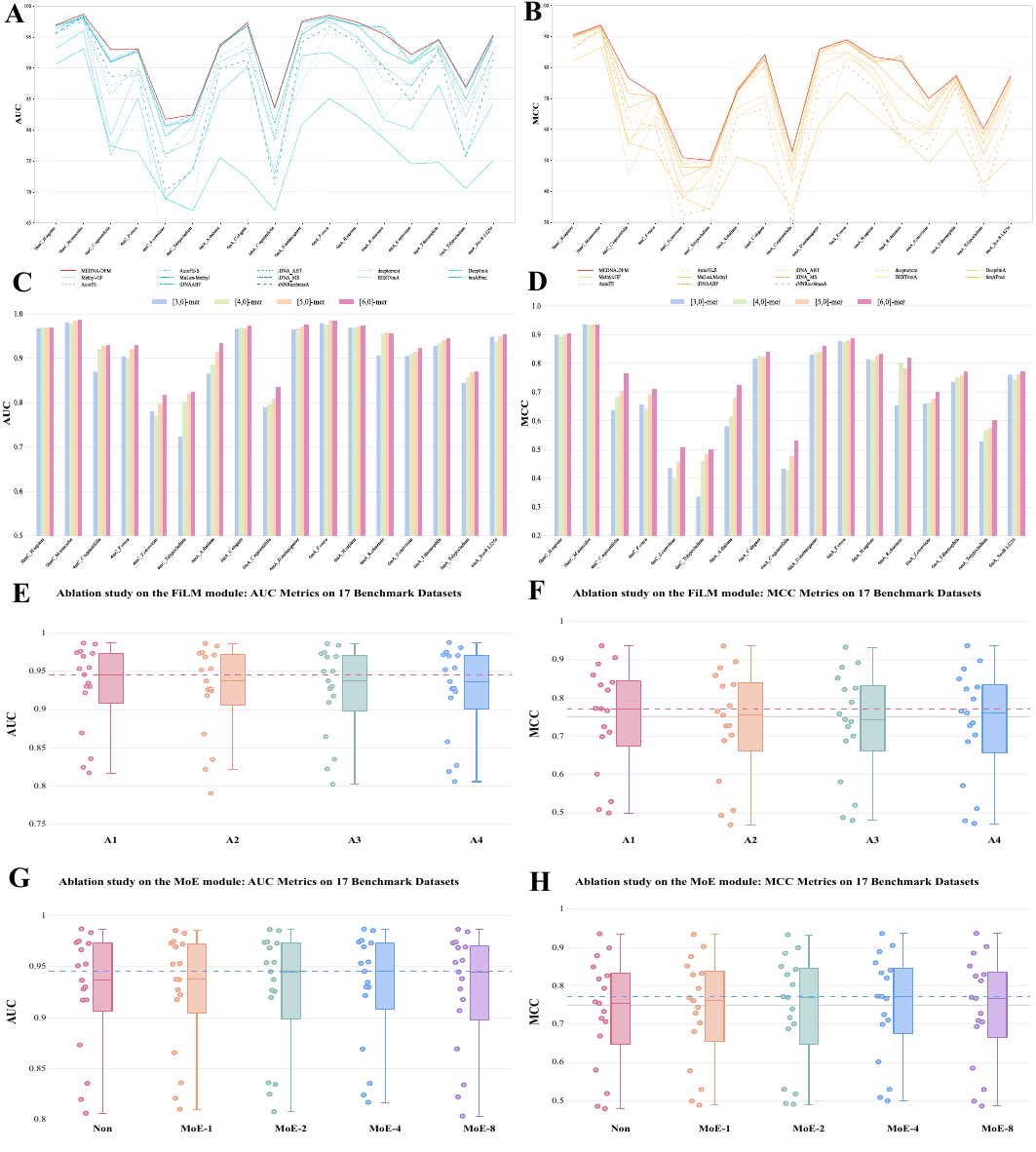}
    \caption{\textbf{Comprehensive performance evaluation and ablation analysis of the MEDNA-DFM.}
    \textbf{(A, B)} Comparison of predictive performance between MEDNA-DFM and representative state-of-the-art methods across 17 benchmark datasets. The line charts illustrate the AUC (A) and MCC (B) scores. \textbf{(C, D)} Impact of token granularity on feature extraction. The bar charts display AUC and MCC metrics for different k-mer tokenization strategies (3, 4, 5, and 6-mer) utilized in the Dual-View DNABERT module. \textbf{(E, F)} Ablation study on the efficacy of the FiLM module. Boxplots summarize the distribution of AUC and MCC values across all datasets for four structural variants: \textbf{A1} (Full Model with Global-to-Local FiLM modulation), \textbf{A2} (Simple Fusion via concatenation), \textbf{A3} (6-mer backbone only), and \textbf{A4} (Reversed FiLM with Local-to-Global modulation). The red dashed line indicates the median performance of the proposed A1 model. \textbf{(G, H)} Analysis of model capacity regarding the MoE module. Boxplots compare the AUC and MCC distributions for the baseline model without MoE (\textbf{Non}) and variants equipped with increasing numbers of experts (\textbf{MoE-1, MoE-2, MoE-4, MoE-8}).}
    \label{fig:result1}
\end{figure}

\paragraph{MEDNA-DFM Establishes a New Benchmark for Multi-Type DNA Methylation Identification with Superior Consistency}
MEDNA-DFM demonstrated highly competitive predictive performance across a comprehensive benchmark of 17 datasets (detailed statistics are summarized in Supplementary Table 1) spanning three DNA modification types (5hmC, 4mC, 6mA)~\cite{lv2020idna}. We benchmarked MEDNA-DFM against a broad spectrum of representative predictors, including iDNA-ABF~\cite{jin2022idna}, Methyl-GP~\cite{xie2025methyl}, AutoFE-Pointer~\cite{feng2025autofe} and others~\cite{zeng2023mulan,yu2021idna,lv2020idna,yu2019snnrice6ma,liu2021deeptorrent,tsukiyama2022bert6ma,li2021deep6ma,pian2020mm}(detailed in Supplementary Table 2). All comparisons were conducted strictly adhering to the original dataset partitions to guarantee a direct and fair assessment. Visual inspection of the performance curves in Fig.\ref{fig:result1}A (AUC) and Fig.\ref{fig:result1}B (MCC) reveals a clear trend: MEDNA-DFM consistently occupies the top-tier boundary across the majority of datasets, demonstrating superior robustness compared to baseline methods. Quantitatively, our model achieved state-of-the-art (SOTA) performance in 6, 6, and 7 out of the 17 datasets for ACC, AUC, and MCC, respectively. For instance, in the 5hmC\_H.sapiens dataset, MEDNA-DFM achieved the highest MCC (90.49\%) and a highly competitive AUC (96.92\%), ranking among the top-performing models. It maintained this consistent performance in the 5hmC\_M.musculus dataset (AUC 98.65\%, MCC 93.58\%). The model's superior predictive power extended to the more challenging 4mC and 6mA modification types. In the 4mC benchmarks, MEDNA-DFM consistently ranked as a leading method, notably achieving the highest MCC (76.64\%) in the C.equisetifolia dataset and a leading AUC (93.00\%) in the F.vesca dataset. For the 10 distinct 6mA datasets, MEDNA-DFM displayed exceptional generalizability. It achieved the highest or second-highest AUC in 7 out of 10 datasets, including C.elegans (97.30\%), D.melanogaster (97.55\%), and H.sapiens (97.40\%). While some methods like AutoFE~\cite{feng2025autofe} or Methyl-GP~\cite{xie2025methyl} showed strong performance on individual datasets, our model's key advantage lies in its exceptional consistency and high-level performance across all datasets.

\paragraph{Ablation Analysis Reveals That Specialized Modules Are Essential for Predictive Superiority}
To systematically deconstruct the structural efficacy of MEDNA-DFM, we evaluated the contribution of each architectural component through extensive ablation studies across diverse biological domains (Fig.\ref{fig:result1}C–H, Supplementary Tables 3, 4, 5).

First, the Dual-View DNABERT module was identified as the cornerstone of feature extraction, with its impact being most pronounced in complex plant genomes. As illustrated in Fig.\ref{fig:result1}C (AUC) and Fig.\ref{fig:result1}D (MCC), a comprehensive evaluation across the 17 datasets reveals a distinct performance hierarchy governed by token granularity. Generally, the 6-mer tokenization strategy consistently yields the most robust results, dominating the top-tier metrics, whereas the 3-mer baseline generally exhibits the lowest predictive capability. This superiority of high-order tokenization is particularly pronounced in complex plant genomes. Specifically,in the 4mC\_C.equisetifolia dataset, our granularity analysis revealed that the 6-mer tokenization strategy is critical for capturing long-range semantic dependencies. It achieved an MCC of 0.7664, overwhelmingly outperforming the 3-mer baseline (MCC 0.6360). This substantial 0.1304 performance margin confirms that coarser-grained tokenization effectively prevents the fragmentation of semantic motifs in high-variability sequences.

Furthermore, the FiLM module proved indispensable for contextual calibration. As depicted in the boxplots of Fig.\ref{fig:result1}E (AUC) and Fig.\ref{fig:result1}F (MCC), the proposed architecture (Variant A1) established a statistically significant performance advantage across the 17 benchmark datasets. The distribution of A1 scores consistently secured the highest median values with minimal variance, indicating superior stability. Interestingly, the MCC landscape (Fig.\ref{fig:result1}F) revealed a crucial directional dependency: while the Reversed FiLM strategy (Variant A4) outperformed the static fusion baselines (Variants A2 and A3), it consistently lagged behind A1. This observation confirms that the "Global-to-Local" modulation flow is optimal for feature refinement, whereas the reverse direction offers only marginal gains. This structural superiority was particularly evident in organisms with dense methylation patterns. Using the 6mA\_T.thermophile dataset, we demonstrated the superiority of our FiLM-based directional architecture (Variant A1) over static fusion baselines. The dynamic modulation of local features by global context yielded an AUC of 0.9452, providing a statistically significant gain of 0.0194 over the direct addition method (Variant A2, AUC 0.9258). This result underscores that simple feature aggregation is insufficient for resolving the intricate epigenetic landscape of thermophilic organisms.

Finally, the Mixture-of-Experts module was validated as effective for balancing model capacity and ensuring stable generalization across diverse biological domains. A comprehensive evaluation across the 17 benchmark datasets (Fig.~\ref{fig:result1}G, H) reveals a clear performance trajectory based on expert allocation. Architectures lacking the MoE routing (Non) or restricted to a single expert (MoE-1) exhibit a significant performance gap compared to the optimal four-expert configuration (MoE-4, indicated by the blue dashed line). While configuring the model with two (MoE-2) or eight (MoE-8) experts narrows this gap, their median performances still slightly lag behind MoE-4. Crucially, the scatter points representing individual datasets are more concentrated within the MoE-4 boxplots. This tighter clustering signifies superior stability and robust prediction consistency across highly variable species, mitigating the extreme performance fluctuations observed in other variants. This global stability translates directly to individual complex genomes. For instance, in the \textit{4mC\_S.cerevisiae} dataset, the optimized MoE-4 configuration emerged as the distinct optimal choice. It achieved an AUC of 0.8170, surpassing the over-parameterized eight-expert variant (MoE-8, AUC 0.8035) by 0.0135. This demonstrates that a coherently scaled multi-expert system effectively mitigates expert redundancy while enhancing the capture of heterogeneous motif patterns. Collectively, these results indicate that the model’s generalizability is not species-specific but stems from a synergistic design capable of adapting to diverse genomic contexts.

\begin{figure}[b]
    \centering
    \includegraphics[width=\textwidth]{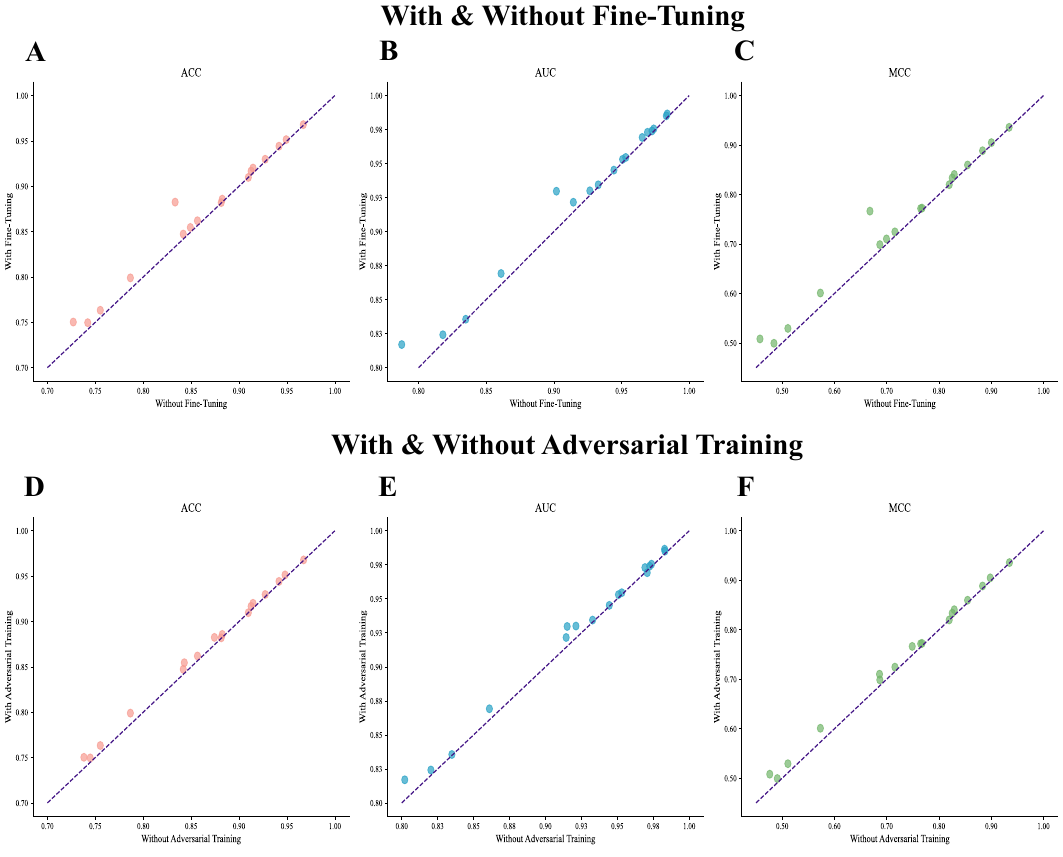}
    \caption{\textbf{Impact of domain-adaptive fine-tuning and adversarial regularization on model robustness.}\textbf{(A,B,C)} Performance comparison with and without fine-tuning. The scatter plots display the ACC (A), AUC (B), and MCC (C) scores across all datasets. The y-axis represents the MEDNA-DFM model utilizing domain-adaptive fine-tuning, while the x-axis represents the model without this phase.\textbf{(D,E,F)} Efficacy of adversarial training. Comparison of ACC (D), AUC (E), and MCC (F) between models trained with (y-axis) and without (x-axis) adversarial regularization. In all plots, each point represents one benchmark dataset. The purple dashed diagonal line ($y=x$) indicates the baseline of identical performance; points located above this line demonstrate that the applied strategy (fine-tuning or adversarial training) yields superior predictive metrics.}
    \label{fig:result2}
\end{figure}
\paragraph{Domain-Adaptive Fine-Tuning and Adversarial Regularization Are Essential for Model Robustness}
Beyond architectural innovations, the specialized training paradigms adopted in this study were identified as pivotal determinants for achieving peak performance (Fig.\ref{fig:result2}, Supplementary Tables 6, 7). Our ablation analysis explicitly highlights the indispensability of domain-adaptive fine-tuning. As illustrated in the scatter plots of Fig.\ref{fig:result2}A–C, the vast majority of data points are situated significantly above the diagonal identity line ($y=x$), visually confirming that the full model consistently outperforms the variant lacking fine-tuning across ACC, AUC, and MCC metrics. The exclusion of this phase resulted in severe performance degradation across all evaluated metrics, particularly in datasets with high epigenetic variability. For instance, in the 4mC\_C.equisetifolia dataset, the model without fine-tuning suffered a drastic collapse in MCC from 0.7664 to 0.6679, representing a relative performance deficit of 9.85\%. This empirical evidence strongly suggests that while pre-trained models provide generalized genomic representations~\cite{ji2021dnabert,zhou2023dnabert}, task-specific adaptation is crucial for aligning these features with the distinct distribution of methylation data~\cite{xie2025methyl}. Complementing this, adversarial training emerged as a vital regularization strategy~\cite{Goodfellow2014ExplainingAH,miyato2017adversarial}. Although the data points in Fig.\ref{fig:result2}D–F cluster closer to the diagonal compared to the fine-tuning plots, they maintain a consistent position above the baseline, indicating a robust, albeit more subtle, performance gain. In the challenging 4mC\_S.cerevisiae dataset, the removal of adversarial perturbations led to a clear MCC drop from 0.5081 to 0.4762 (-0.0319), while a similar trend was observed in 4mC\_F.vesca (MCC decrease from 0.7102 to 0.6862). These results indicate that adversarial training plays a nuanced yet contributory role in smoothing the decision boundary, thereby enhancing the model's resilience against input variations. Consequently, the predictive superiority of MEDNA-DFM is not merely a product of its static architecture but is significantly reinforced by the dynamic interplay of these advanced training strategies.

\begin{figure}[p]
    \centering
    \includegraphics[width=1.0\textwidth]{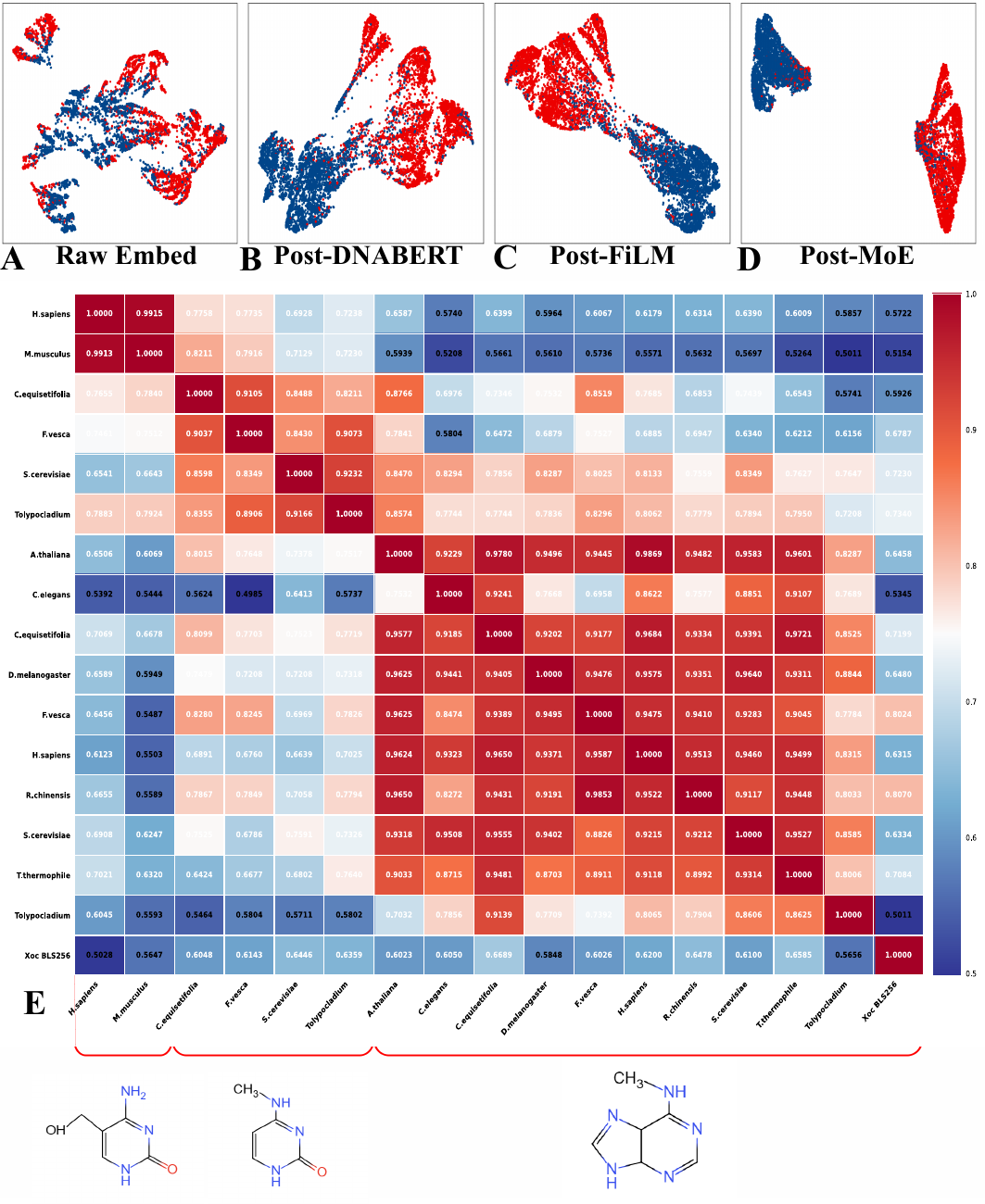}
    \caption{\textbf{Visualization of internal representation dynamics and cross-species generalization capabilities.}\textbf{(A--D)} Trajectory of feature space evolution via UMAP visualization. The scatter plots illustrate the distribution of samples at four key processing stages: \textbf{(A)} Raw Embeddings, \textbf{(B)} Post-DNABERT, \textbf{(C)} Post-FiLM, and \textbf{(D)} Post-MoE. Blue points represent positive samples, while yellow points represent negative samples. \textbf{(E)} Cross-dataset heatmap. The matrix depicts the pairwise evaluation where rows represent source datasets (training) and columns represent target datasets (testing). Color intensity reflects the relative transferability, calculated as the ACC on the target set normalized by the source-on-source baseline. The chemical structures at the bottom correspond to the three DNA methylation categories: 5-hydroxymethylcytosine (5hmC), N4-methylcytosine (4mC), and N6-methyladenine (6mA).}
    \label{fig:result3}  
\end{figure}
\paragraph{MEDNA-DFM Achieves Effective Methylation Distinction and Captures Conserved Methylation Patterns Across Species}
We investigated the internal representation dynamics of MEDNA-DFM to validate its ability to strictly distinguish methylation states and uncover biologically intrinsic patterns. First, the model demonstrates a progressive capability for methylation discrimination, effectively separating positive (red) from negative (blue) samples. Tracking the feature space via UMAP visualization~\cite{Healy2024} reveals that the decision boundary between methylated and non-methylated sequences becomes increasingly distinct at each key processing stage (Fig.\ref{fig:result3}A–D; see Supplementary Fig.S2 for all 17 datasets). Specifically, the separability improves markedly from the initial fine-tuning phase, through the context-guided modulation by the FiLM module, and crystallizes finally after integration by the MoE module (Fig.\ref{fig:result3}D). This stepwise refinement visually corroborates the ablation studies, confirming that each architectural component actively contributes to distilling discriminative biological signals from the raw sequence. Second, Motivated by the observation that identical methylation types share intrinsic sequence patterns (e.g., the conserved CpG motifs for 5hmC or [G/C]AGG[C/T] for 6mA across species)~\cite{fu2015n6,greer2015dna,tahiliani2009conversion}, we adopted a multi-species integration strategy. The cross-dataset evaluation (Fig.\ref{fig:result3}E, the raw and normalized cross-species ACC matrices are detailed in Supplementary Tables 8 and 9, respectively) confirms the effectiveness of this approach. The heatmap displays a distinct block-diagonal structure, where species belonging to the same modification type (visualized with their corresponding chemical structures for 5hmC, 4mC, and 6mA) cluster tightly together with high prediction correlations. This indicates that MEDNA-DFM has successfully disentangled the fundamental, modification-specific semantic patterns from species-specific genomic background noise. Instead of memorizing species identities, the model focuses on the evolutionarily conserved biochemical signatures, thereby achieving robust generalization within each modification family while maintaining strict boundaries between distinct chemical modifications.

\begin{figure}[p]
    \centering
    \includegraphics[width=\textwidth]{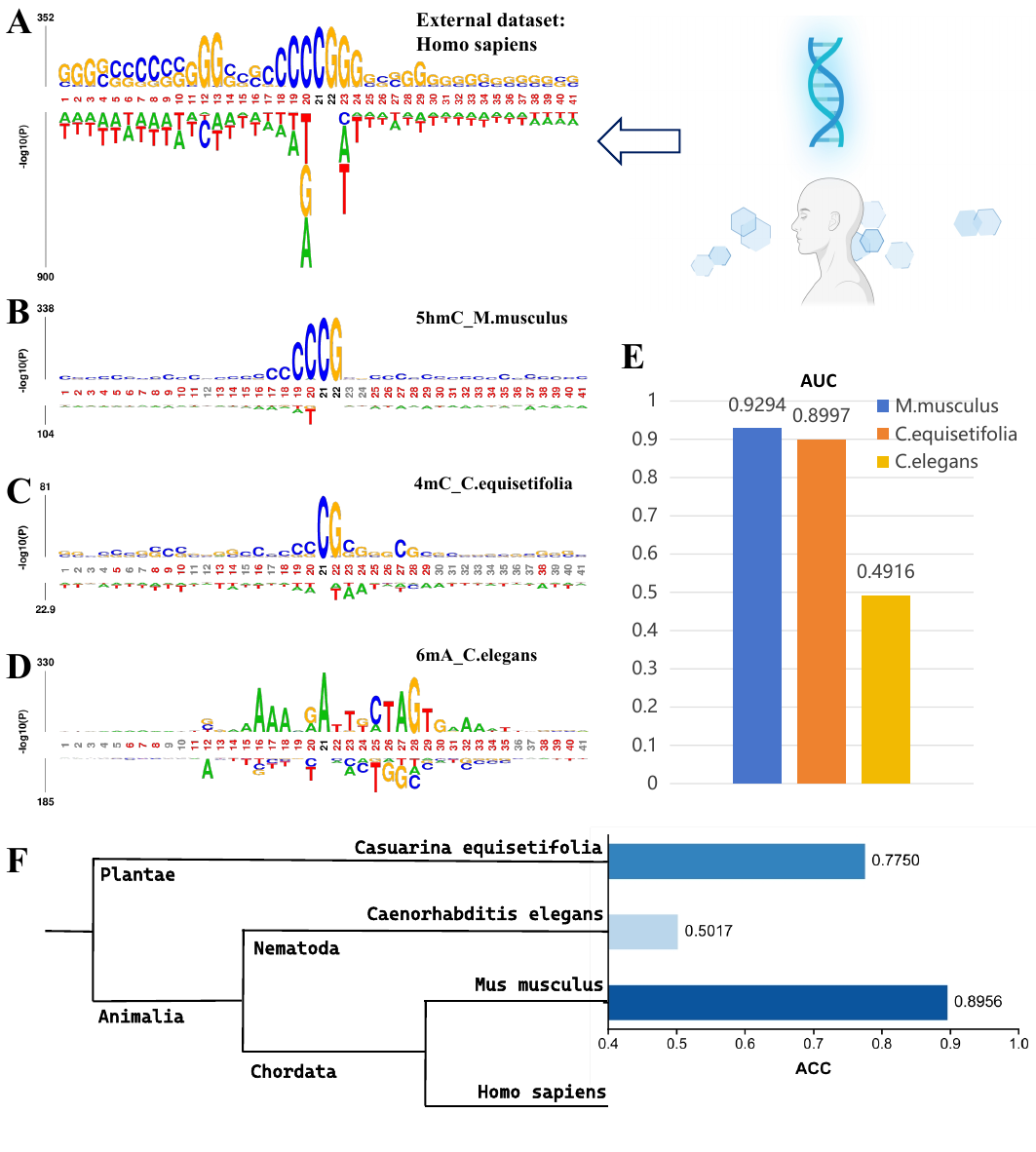} 
    \caption{\textbf{Mechanism of motif-driven generalization across evolutionary boundaries.}
    \textbf{(A--D)} Comparative motif analysis using kpLogo illustrating the sequence preferences of the target and source datasets. \textbf{(A)} The external independent validation set \textit{Homo sapiens} (5mC). \textbf{(B--D)} The source datasets used for training: \textbf{(B)} \textit{M. musculus} (5hmC), \textbf{(C)} \textit{C. equisetifolia} (4mC), \textbf{(D)} \textit{C. elegans} (6mA).
    \textbf{(E)} AUC of the three source models directly applied to the external \textit{H. sapiens} dataset.
    \textbf{(F)} Contrast between phylogenetic distance and predictive accuracy (ACC). The phylogenetic tree shows the evolutionary proximity of the species to \textit{H. sapiens}, while the bar chart (right) reveals that the evolutionarily distant plant model (\textit{C. equisetifolia}) significantly outperforms the closely related nematode model (\textit{C. elegans}).}
    \label{fig:result4}
\end{figure}

\paragraph{External Validation Reveals Conserved Motif-Driven Generalization in MEDNA-DFM}
To rigorously evaluate the biological generalization capability of MEDNA-DFM, we employed a cross-modification design to verify whether the model genuinely captures universal epigenetic grammars rather than merely overfitting to the enzymatic signatures of a single methylation type. We performed a systematic external validation using an independent \textit{Homo sapiens} 5mC dataset against models pre-trained on all 17 benchmark datasets (detailed in Supplementary Table 9). We selected three representative models—\textit{Mus musculus} (5hmC), \textit{Casuarina equisetifolia} (4mC), and \textit{Caenorhabditis elegans} (6mA)—to further characterize the relationship between predictive performance and evolutionary distance (Fig.~\ref{fig:result4}). As illustrated in Fig.~\ref{fig:result4}E and Fig.~\ref{fig:result4}F , the model trained on \textit{M. musculus} achieved the highest generalization performance (AUC = 0.9294, ACC = 0.8956). This is expected, given the close phylogenetic relationship between humans and mice (both Chordata). However, a counter-intuitive phenomenon emerged when comparing the other two models. As shown in the phylogenetic tree in Fig.~\ref{fig:result4}F, \textit{C. elegans} (Animalia) is evolutionarily closer to \textit{H. sapiens} than \textit{C. equisetifolia (Plantae)}. Despite this, the \textit{C. equisetifolia} model significantly outperformed the \textit{C. elegans} model, achieving an AUC of 0.8997 and ACC of 0.7750, whereas the \textit{C. elegans} model failed to generalize effectively (AUC = 0.4916, ACC = 0.5017). The kpLogo~\cite{wu2017kplogo} analysis reveals the underlying mechanism(Fig.~\ref{fig:result4}A–D; comprehensive motif profiles for all 17 benchmark datasets are provided in Supplementary Fig.S3): The external \textit{H. sapiens} 5mC dataset exhibits a strong enrichment of CG dinucleotides(Fig.~\ref{fig:result4}A), a hallmark of mammalian methylation~\cite{lister2009human}. Similarly, both the \textit{M. musculus} (5hmC) and \textit{C. equisetifolia} (4mC) datasets are characterized by GC-rich motifs(Fig.~\ref{fig:result4}B, C), consistent with the high GC content often observed in methylated regions of plants~\cite{law2010establishing}. This structural similarity explains why the "Plant" model transfers well to Human data. In contrast, the \textit{C. elegans} 6mA dataset is dominated by A-rich motifs (e.g., ApT or TpA context) and lacks CG enrichment(Fig.~\ref{fig:result4}D)~\cite{greer2015dna}. Consequently, the model trained on this data looks for features that do not exist in the Human 5mC context, leading to near-random prediction performance. These results demonstrate that MEDNA-DFM successfully captures intrinsic biochemical rules—specifically, the preference for GC-rich versus AT-rich environments—across species boundaries. The model's ability to prioritize sequence syntax over species labels and methylation types validates its potential as a generalizable tool for exploring methylation mechanisms in uncharacterized genomes.

\begin{figure}[b!]
    \centering
    \includegraphics[width=\textwidth]{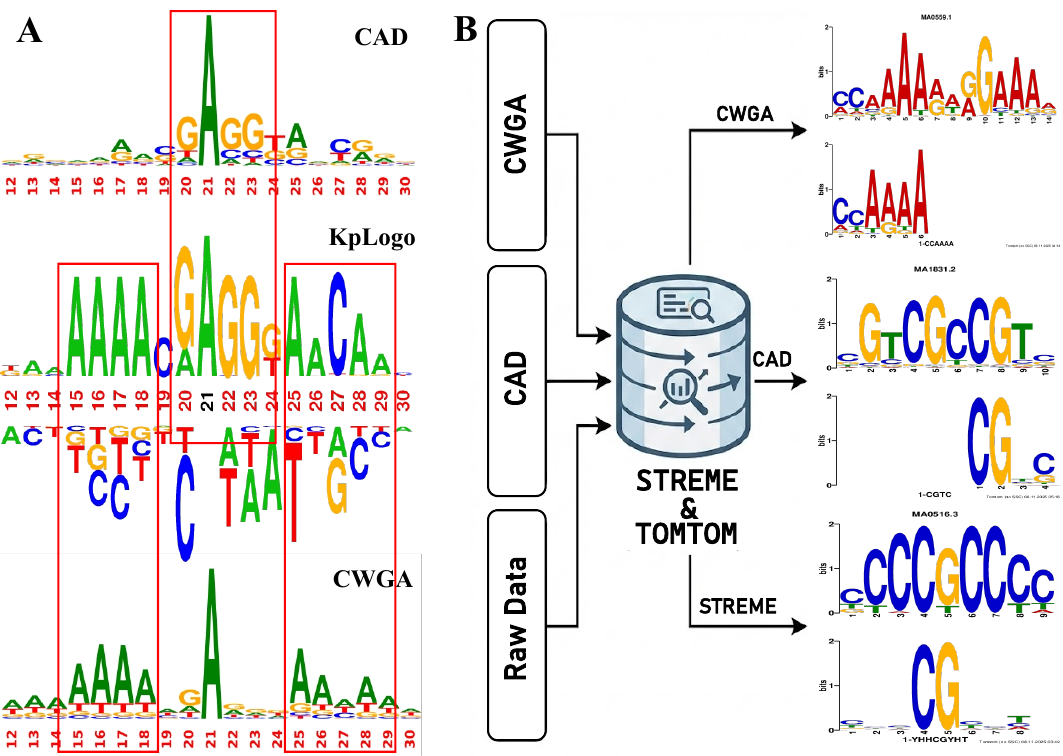} 
    \caption{\textbf{Signal disentanglement and high-fidelity motif purification via MEDNA-DFM.} 
    \textbf{(A)} Comparison of sequence logo landscapes for the \textit{6mA\_D.melanogaster} dataset. The traditional KpLogo analysis effectively represents a composite union of these two independent feature subsets.
    \textbf{(B)} Analytical workflow for model-guided motif discovery and validation. Sequence features extracted from CAD and CWGA methods, along with raw data, are processed through STREME. The resulting motifs are validated via TOMTOM to identify statistically significant matches to known transcription factor binding sites.}
    \label{fig:result5}
\end{figure}

\paragraph{Signal Disentanglement and Purification Enables Highly Reliable Motif Mining} 
First, our analysis demonstrates that MEDNA-DFM functions as a statistical signal disentangler, effectively deconstructing complex biological patterns into distinct functional components. As observed in the \textit{6mA\_D.melanogaster} dataset (Fig.~\ref{fig:result5}A), the model separates the local biochemical context from distal sequence dependencies. Specifically, the DNABERT (CAD) view concentrates intensely on the central adenine 'A' and its immediate neighbors (guanine 'G' and cytosine 'C'), capturing the fine-grained local modification environment. In contrast, the FiLM (CWGA) view, while acknowledging the central site, allocates significant decision weights to distal flanking adenines. Interestingly, the complete sequence motif landscape generated by the traditional KpLogo analysis~\cite{wu2017kplogo} could be almost precisely viewed as the union of the two independent feature subsets derived from CAD and CWGA(Fig.~\ref{fig:result5}A).

Furthermore, this signal disentanglement capability directly translates into superior motif mining performance, enabling the purification of core features and the unmasking of auxiliary elements submerged in raw sequence data. We validated this by comparing our model-guided discovery against standard STREME~\cite{bailey2021streme} analysis on raw sequences using the \textit{5hmC\_H.sapiens} dataset (Fig.~\ref{fig:result5}B). While direct application of STREME to raw data identified a statistically significant CG-enriched motif ($E\text{-value}=5.4\times10^{-24}$), its biological relevance was diluted by background noise, as evidenced by a relatively low-confidence match to the SP2 transcription factor (MA0516.3; $p=1.85\times10^{-4}$). In stark contrast, our method acted as a high-fidelity filter: the motif extracted via the DNABERT view (CAD) achieved a markedly higher statistical match to the EGR1 binding site (MA1831.2; $p=2.25\times10^{-5}$; comprehensive CAD purification metrics and downstream TOMTOM alignments are detailed in Supplementary Table 10 and Supplementary Fig.S4). Crucially, the model also recovered functional signals completely overlooked by traditional analysis. The FiLM view (CWGA) successfully unmasked a distinct auxiliary motif, 1-CCAAAA, which corresponds to the PI transcription factor (MA0559.1) (Supplementary Table 11 and Supplementary Fig.S5). This motif was completely submerged in the STREME results from the raw data.

To quantitatively validate motif reliability, we compared the statistical significance ($p$-values) between our methods against raw data analysis (STREME) and interpretable models Methyl-GP~\cite{xie2025methyl} and iDNA-ABF~\cite{jin2022idna}. As shown in Table 1 and 2, our method consistently identified motifs with substantially higher confidence levels.
\vspace{-1.5em}
\begin{table}[hbp]
  \centering
  \small
  \caption{Comparison of $p$-values for 6mA in \textit{A.thaliana} across different methods. Baseline results for Methyl-GP and its corresponding STREME analysis are adopted from Xie et al.~\cite{xie2025methyl}.}
  \label{tab:table1}
  \setlength{\tabcolsep}{3.0pt}
  \renewcommand{\arraystretch}{1.2}  
  \begin{tabular}{@{}lccccccc@{}}  
    \toprule  
    & STREME 
    & \makecell{Methyl-GP\\(3mer)}  
    & \makecell{Methyl-GP\\(4mer)}
    & \makecell{Methyl-GP\\(5mer)}
    & \makecell{Methyl-GP\\(6mer)}
    & CAD 
    & CWGA \\
    \midrule  
    & $4.1 \times 10^{-4}$ 
    & $1.25 \times 10^{-2}$ 
    & $2.25 \times 10^{-3}$ 
    & $2.18 \times 10^{-2}$ 
    & $4.66 \times 10^{-2}$ 
    & $\mathbf{1.37 \times 10^{-5}}$
    & $6.16 \times 10^{-5}$ \\
    \bottomrule  
  \end{tabular}
\end{table}
\vspace{-3.0em}
\begin{table}[hbp]
  \centering
  \caption{Comparison of $p$-values across different methylation types and species. Baseline results for iDNA-ABF and its corresponding STREME analysis are adopted from Jin et al.~\cite{jin2022idna}.}
  \label{tab:table2}
  \renewcommand{\arraystretch}{1.5}
  \begin{tabular}{lccc}
    \hline
     & 4mC\_Tolypocladium & 5hmC\_H.sapiens & 6mA\_C.equisetifolia \\
    \hline
    STREME & $5.0 \times 10^{-2}$ & $9.3 \times 10^{-1}$ & $2.7 \times 10^{-2}$ \\
    iDNA\_ABF & $2.83 \times 10^{-2}$ & $5.47 \times 10^{-2}$ & $2.04 \times 10^{-3}$ \\
    CAD & $\mathbf{7.85 \times 10^{-4}}$ & $\mathbf{2.25 \times 10^{-5}}$ & $\mathbf{5.54 \times 10^{-4}}$ \\
    CWGA & $1.77 \times 10^{-3}$ & $1.45 \times 10^{-4}$ & $9.68 \times 10^{-4}$ \\
    \hline
  \end{tabular}
\end{table}
\vspace{-1.8em}
In the $6mA\_A.thaliana$ dataset (Table 1), the multi-scale k-mer analysis of Methyl-GP yielded $p$-values ranging from $1.25 \times 10^{-2}$ to $4.66 \times 10^{-2}$, indicating only marginal significance. Even the direct STREME analysis ($p=4.1 \times 10^{-4}$) outperformed Methyl-GP. However, our CAD view achieved a decisive significance level of $1.37 \times 10^{-5}$, representing an improvement of nearly two orders of magnitude over the best Methyl-GP configuration. This superiority was further corroborated in comparisons with iDNA-ABF (Table 2). For instance, in the complex $5hmC\_H.sapiens$ dataset, the motif extracted by iDNA-ABF ($p=5.47 \times 10^{-2}$) and raw STREME ($p=9.3 \times 10^{-1}$) failed to pass stringent significance thresholds. In stark contrast, our model successfully purified the signal, identifying a core motif with a highly significant $p$-value of $2.25 \times 10^{-5}$. Similar trends were observed in $4mC\_Tolypocladium$ and $6mA\_C.equisetifolia$, where both CAD and CWGA views consistently surpassed the baselines. These results confirm that MEDNA-DFM acts as a robust statistical amplifier, effectively suppressing background noise to reveal highly reliable and stable biological signals that remain obscured or fluctuate in competing models.

\begin{figure}[b!]
    \centering
    \includegraphics[width=\textwidth]{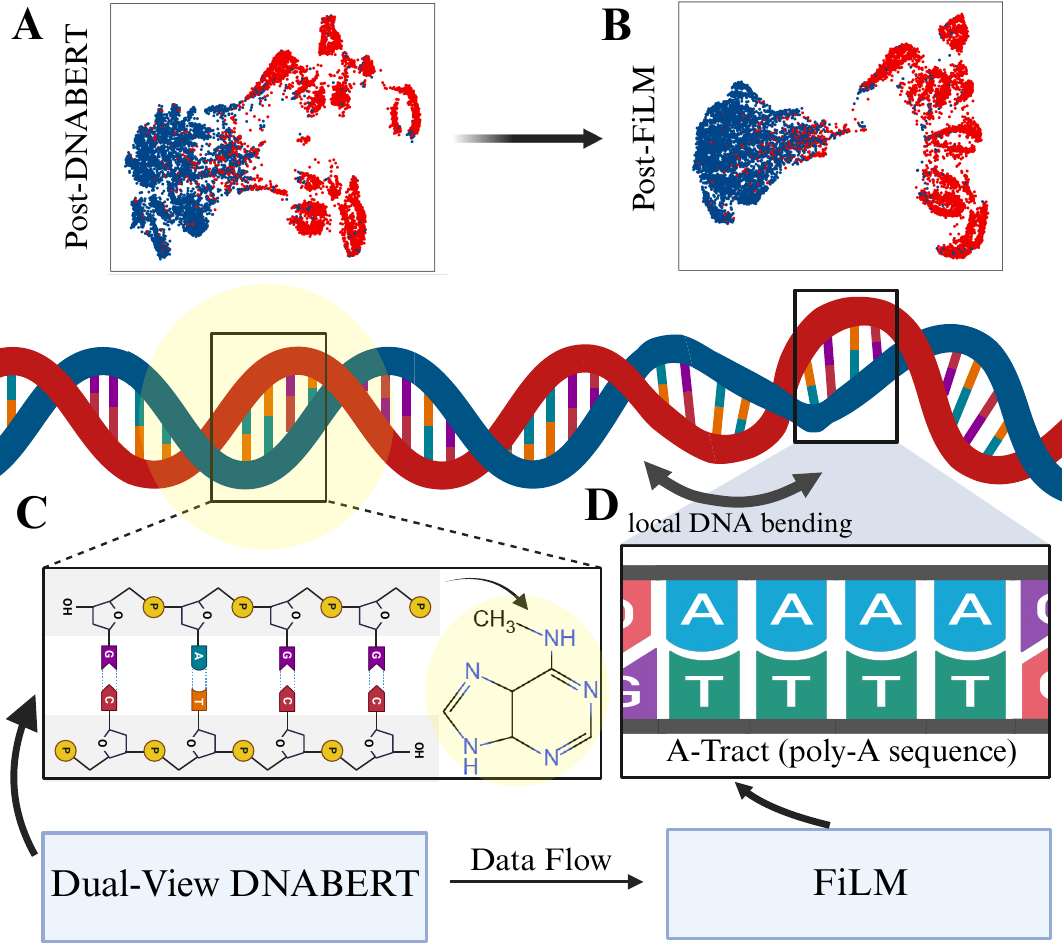} 
    \caption{\textbf{Computational deconstruction of the sequence-structure dependency in \textit{Drosophila} 6mA regulation.} 
    The schematic illustrates the progressive purification of biological signals as data flows through the MEDNA-DFM architecture. 
    \textbf{(A)} UMAP visualization of the feature space at the Post-DNABERT stage. While initial discrimination is achieved, positive samples (red) remain dispersed into disparate sub-clusters. 
    \textbf{(B)} UMAP visualization at the Post-FiLM stage. This stage effectively draws the previously divergent sub-clusters together. 
    \textbf{(C)} Corresponding to (A), the Dual-View DNABERT module captures the local \textbf{GAGG core motif}
    \textbf{(D)} Corresponding to (B), the FiLM module captures the A-Tract, which induces \textbf{local DNA bending}.}
    \label{fig:result6}
\end{figure}
\paragraph{Case Study: Computational Deconstruction of Drosophila 6mA Reveals a Sequence-Structure Dependency}
To investigate the underlying patterns of \textit{Drosophila} 6mA regulation, we analyzed the internal representations of the MEDNA-DFM model using UMAP dimensionality reduction (Fig.~\ref{fig:result6}). This analysis revealed a distinct functional dichotomy between the model's modules. First, the DNABERT module precisely captured the GAGG core motif centered on the modification site (Fig.~\ref{fig:result6}C), consistent with the canonical consensus identified in Drosophila~\cite{zhang2015n6}. While this feature provided an initial separation of methylated (positive) and non-methylated (negative) samples, the UMAP visualization indicated that positive samples were dispersed into multiple disparate sub-clusters (Fig.~\ref{fig:result6}A). This divergence suggests that while the GAGG motif is a necessary anchor, the sequence contexts surrounding it are highly variable, preventing a unified representation based on sequence identity alone. Crucially, the FiLM module identified a secondary, stable signal: a dependency on an A-Tract (poly-A sequence) (Fig.~\ref{fig:result6}D). Mapping this sequence to DNA shape parameters~\cite{rohs2009role} confirms that it induces a significant narrow minor groove and local DNA bending~\cite{nelson1987structure}. When this structural feature was incorporated, the previously divergent sub-clusters in the UMAP space were effectively drawn together, maximizing the global separation between positive and negative classes (Fig.~\ref{fig:result6}B). This computational observation demonstrates that the \textit{Drosophila} 6mA signal is not defined by a monolithic sequence, but rather by a composite signature consisting of a variable GAGG sequence core spatially coupled with a conserved upstream structural element, a mechanism reminiscent of the "shape-sequence" dual recognition mode observed in transcription factors~\cite{slattery2014absence}.

\paragraph{\textit{In silico} Mutagenesis Validates the Sequence-Structure Dependency and Algorithm Faithfulness}
To quantitatively validate the sequence-structure dependency observed in the \textit{Drosophila melanogaster} case study, and to simultaneously evaluate the interpretative faithfulness of our proposed CAD and CWGA algorithms, we performed a targeted \textit{in silico} mutagenesis experiment (Fig.~\ref{fig:result7}; comprehensive quantitative results are detailed in Supplementary Table 12)~\cite{alipanahi2015predicting,zhou2015predicting}. We systematically perturbed the core sequence motif (GAGG) and the structural element (A-tract) in the positive samples of the test set. To rigorously eliminate the confounding effects of altered background distributions, the nucleotide frequencies were strictly preserved during perturbations~\cite{whalen2022navigating,ghandi2014enhanced}. Specifically, the core GAGG motif was mutated to CTCC, and consecutive A-tracts(AAAA) were replaced with alternating TATA sequences (Fig.~\ref{fig:result7}A). Negative samples were kept intact to serve as an internal control. The experimental results present a compelling and rigorously controlled evidence chain. As illustrated in Fig.\ref{fig:result7}B, across all perturbation groups, the specificity (SP) for negative samples remained exceptionally constant at 0.9299. This demonstrates that our mutation strategy successfully maintained the background distribution, ensuring that any observed performance degradation is entirely attributable to the ablation of true positive methylation signals rather than a systemic model collapse. In contrast, the sensitivity (SN) exhibited a stepwise decline, directly corroborating the synergistic regulation mechanism. Independent disruption of either the core motif (Mut-GAGG) or the upstream structural element (Mut-A-tract) resulted in a substantial failure to recognize true 6mA sequences, with SN dropping from 0.9298 (Wild-type) to 0.8810 and 0.8574, respectively. This strongly indicates that even with an intact GAGG core, the absence of the cooperative A-tract structural element impairs the model's predictive capability. Furthermore, the global performance metrics evaluation (Fig.~\ref{fig:result7}C)  revealed that the simultaneous ablation of both elements (Mut-Both) yielded the severe performance degradation. The overall accuracy (ACC) dropped from 0.9298 to 0.8622, AUC decreased from 0.9755 to 0.9441, and the Matthews correlation coefficient (MCC), a metric highly sensitive to classification errors, plummeted from 0.8597 to 0.7311. Importantly, the precise targeting of these perturbations was strictly guided by the specific features extracted via our CAD and CWGA algorithms. The fact that disrupting these algorithm-identified regions triggers such a profound loss in predictive power provides direct computational evidence that CAD and CWGA successfully isolate the true, causal features driving the model's decisions, rather than capturing statistical artifacts. Together, these results serve a dual purpose: they not only validate that the MEDNA-DFM model's decision-making is firmly grounded in the authentic biological sequence-structure dependency of 6mA regulation, but also demonstrate the high faithfulness and utility of our interpretation algorithms.
\begin{figure}[t!]
    \centering
    \includegraphics[width=\textwidth]{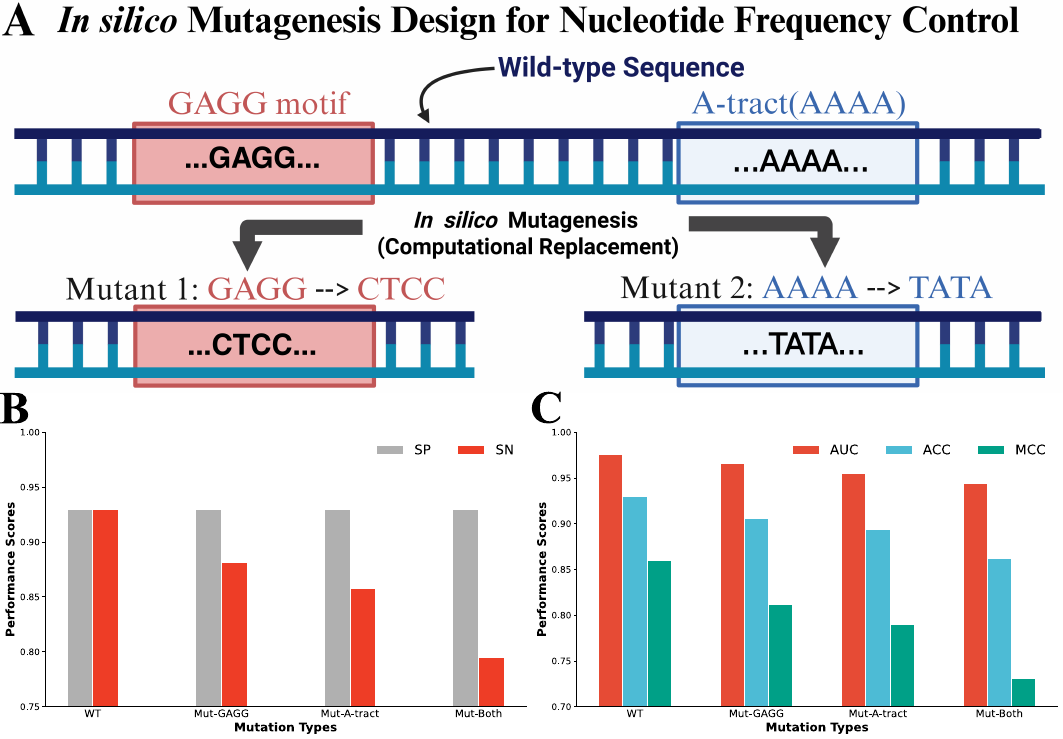} 
    \caption{\textbf{Targeted \textit{in silico} mutagenesis reveals the sequence-structure dependency and validates the faithfulness of interpretation algorithms.} The top schematic \textbf{(A)} illustrates the computational perturbation strategy applied to the positive test samples. The GAGG core motif and the upstream A-tract (AAAA) were specifically mutated to CTCC and alternating TATA, respectively. Negative samples were kept intact as an internal control. \textbf{(B)} Comparison of Sensitivity (SN) and Specificity (SP) across the wild-type (WT) and three mutated conditions (Mut-GAGG, Mut-A-tract, and Mut-Both). \textbf{(C)} Evaluation of global predictive performance metrics, including Accuracy (ACC), Area Under the Curve (AUC), and Matthews Correlation Coefficient (MCC).}
    \label{fig:result7}
\end{figure}
\section{Discussion}\label{sec12}
In this study, we first validated MEDNA-DFM's predictive superiority across 17 benchmarks, with ablation studies confirming the critical efficacy of architecture. Extending beyond standard metrics, our external validation revealed that MEDNA-DFM prioritizes intrinsic sequence syntax over species and modification labels, demonstrating a motif-driven generalization that transcends both evolutionary and chemical boundaries. Unlike previous studies that stopped at performance benchmarking or superficial attention visualization \cite{jin2022idna, xie2025methyl}, we advanced to a deep dissection of the model's internal decision-making mechanism, which directly inspired the development of our CAD and CWGA algorithms. By applying these algorithms for signal purification, we distilled high-confidence motifs with statistical significance two to three orders of magnitude higher than those identified by previous attention-based methods (e.g., iDNA-ABF) or standard motif scanners (e.g., STREME).

Most notably, our computational deconstruction offers a novel mechanistic perspective on Drosophila 6mA regulation, moving beyond simple sequence pattern matching. Based on the computational observation that sequence (GAGG) and structure (A-Tract) features are spatially coupled, we rigorously validated this sequence-structure dependency via targeted in silico mutagenesis. The profound performance degradation observed upon ablating these specific regions confirmed their cooperative necessity, while simultaneously demonstrating the high faithfulness of our CAD and CWGA algorithms in isolating true causal features. Driven by this validated computational dependency, we propose a "Sequence-Shape Synergy" hypothesis for Drosophila 6mA regulation. We postulate that the recruitment of the 6mA methyltransferase (Writer) complex is not a single-step event but a sequential process. The complex may first anchor to the GAGG motif via sequence-specific DNA binding domains, followed by a conformation confirmation step triggered by the narrow minor groove and DNA bending induced by the upstream A-Tract. This "structural sensor" mechanism ensures that methylation occurs only when both the correct sequence and the permissive local topology are present.

Crucially, this binary regulation hypothesis offers a unifying explanation for the persistent controversies in Drosophila 6mA research. First, it redefines the previously reported "AG-rich" consensus (e.g., AGAGGAG)~\cite{Yao2018active}. We argue that this elongated motif is not a monolithic recognition sequence, but rather a statistical amalgam of two functionally distinct signals—the GAGG core and the A-rich structural element—which our model successfully decouples. Second, it resolves the paradox of "extremely low 6mA levels"~\cite{boulet2023adenine, o2019sources}. While GAGG motifs are abundant across the genome, the probability of a GAGG motif co-occurring with a structural A-Tract in the exact spatial configuration required for the "AND-gate" activation is exceedingly low. This aligns with the principle that multicellular eukaryotes employ high "contextual redundancy" (or syntax) to ensure regulatory precision~\cite{yang2025interpretablestructuralsemanticdecodingreveals}. Thus, the observed scarcity of 6mA likely reflects this exquisite regulatory stringency rather than experimental noise or contamination.

Despite these promising insights, we acknowledge certain limitations regarding the scope of our current study. Methodologically, the validity and faithfulness of CAD and CWGA have been robustly established through external biological validation (including qualitative and quantitative comparisons with authoritative tools like kpLogo and STREME, as well as baseline models) and \textit{in silico} mutagenesis. However, in terms of biological scope, MEDNA-DFM is primarily designed to decipher DNA sequence syntax. We recognize that DNA methylation in vivo is also influenced by dynamic spatial contexts, such as chromatin accessibility~\cite{klemm2019chromatin} and nucleosome positioning~\cite{struhl2013determinants}. However, incorporating these multi-dimensional features into a unified model is currently constrained by the limited availability and inherent noise of high-resolution spatial datasets across non-model species. Therefore, rather than attempting to capture the full in vivo complexity, our framework serves as a reliable sequence-level pre-screening mechanism. By efficiently narrowing down the target search space, CAD and CWGA provide a computational baseline that can help prioritize candidates, thereby reducing experimental costs and accelerating downstream biological discoveries and clinical biomarker identification.

\section{Methods}\label{sec11}

\subsection{Datasets}
\paragraph{Benchmark Datasets}
We reuse the exact data sources and original train/test partitions from the iDNA-MS benchmark, without any alteration~\cite{lv2020idna}. This benchmark, established by Lv et al., comprises 17 independent datasets covering multiple species, various DNA methylation types (4mC, 6mA, and 5hmC), and a uniform sequence length of 41 bp. Per the iDNA-MS benchmark definition, each sample is labeled as either positive (modification site) or negative (non-modification site). Detailed statistics for these 17 datasets, including species, modification type, and the number of positive and negative samples in the training and test sets, are exhaustively listed in Supplementary Table 1.

\paragraph{External Datasets}
For independent validation, we curated a balanced human dataset ($N=27,900$) from the BERT-5mC repository~\cite{wang2023bert} by retaining all positive samples and under-sampling the top-ranked negative sequences to achieve a 1:1 ratio. Evaluation was performed via direct zero-shot inference using the frozen pre-trained MEDNA-DFM, strictly excluding any fine-tuning or data leakage.

\subsection{The MEDNA-DFM Architecture}
\begin{figure}[ht]
    \centering
    \includegraphics[width=1.0\textwidth]{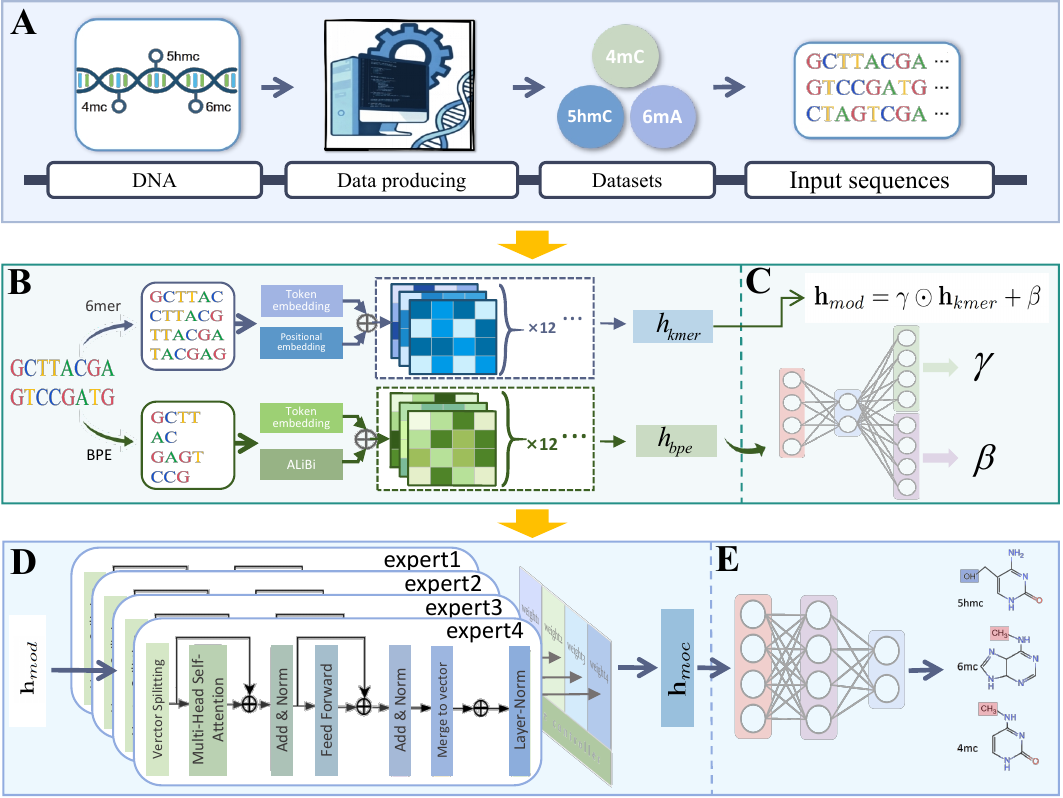}
    \caption{\textbf{Architecture of the MEDNA-DFM.} \textbf{(A)} Overview of the benchmark data processing flow. \textbf{(B)} The Dual-View DNABERT module, featuring two parallel encoders. \textbf{(C)} The FiLM (Feature-wise Linear Modulation) module. \textbf{(D)} The Mixture of Experts (MoE) module. \textbf{(E)} The final Classification Module, which maps the aggregated high-dimensional features to binary prediction logits.}
    \label{fig:DFM}  
\end{figure}

\paragraph{Overview}
MEDNA-DFM (Dual-View FiLM-MoE Model) is a deep learning architecture designed for the binary classification of DNA methylation (Fig.\ref{fig:DFM}). The architecture comprises four core components: (i) a Dual-View DNABERT Feature Extraction module, (ii) a FiLM module, (iii) an MoE module, and (iv) a Final Prediction Layer.

\paragraph{Dual-View DNABERT Module}
The model employs two parallel encoders to process the input DNA sequence $S$ (length $L=41$).
(1) DNABERT-6mer: The sequence is tokenized into overlapping 6-mers and processed by DNABERT. The pooler output—derived from the special [CLS] token via a dense layer with Tanh activation—is serves as the $k$-mer representation $\mathbf{h}_{kmer}$.
(2) DNABERT2: The same sequence $S$ is tokenized via Byte Pair Encoding (BPE) and processed by DNABERT2. The final hidden states of all BPE tokens are subjected to mean pooling to generate the contextual representation $\mathbf{h}_{bpe}$.

\paragraph{FiLM Module}
To synthesize the two representations, we implement a Feature-wise Linear Modulation (FiLM) mechanism. An auxiliary Multilayer Perceptron (MLP) maps the contextual representation $\mathbf{h}_{bpe}$ to affine modulation parameters. The MLP consists of two linear layers with an intermediate ReLU activation, formulated as:$$[\gamma, \beta] = \text{Linear}_{2}(\text{ReLU}(\text{Linear}_{1}(\mathbf{h}_{bpe})))$$The resulting output vector ($\in \mathbb{R}^{2d}$) is partitioned along the feature dimension into a scale vector $\gamma \in \mathbb{R}^{d}$ and a shift vector $\beta \in \mathbb{R}^{d}$ ($d=768$). These coefficients constitute an element-wise affine transformation to the $k$-mer representation $\mathbf{h}_{kmer}$:$$\mathbf{h}_{mod} = \gamma \odot \mathbf{h}_{kmer} + \beta$$where $\odot$ denotes the Hadamard product.

\paragraph{Mixture of Experts (MoE) Module}

The MoE module aggregates features from $N=4$ experts, conditioned on $\mathbf{h}_{bpe}$. The process consists of three stages:

\textbf{Gating Network.} The gating network computes a probability distribution $\mathbf{w} \in \mathbb{R}^N$ to weight the experts. It projects the input $\mathbf{h}_{bpe}$ followed by a Softmax normalization:$$\mathbf{w} = \text{Softmax}(\text{Linear}_{N}(\mathbf{h}_{bpe}))$$

\textbf{Expert Architecture.} The module comprises $N$ isomorphic experts, each processing the modulated representation $\mathbf{h}_{mod}$. The input vector ($d=768$) is reshaped into a sequence of $m=4$ segments (dimension $192$ each). This sequence is encoded by a single-layer Transformer ($n_{\text{head}}=4$, $d_{\text{ff}}=256$, with GELU activation). The output is flattened and fused with the original input via a residual connection and Layer Normalization:$$\mathbf{h}_{expert\_i} = \text{LayerNorm}(\mathbf{h}_{mod} + \text{Flatten}(\text{Transformer}(\text{Reshape}(\mathbf{h}_{mod}))))$$

\textbf{Feature Aggregation.} The final representation is derived by aggregating the expert outputs according to weights $\mathbf{w}$, followed by a final linear projection:$$\mathbf{h}_{moe} = \text{Linear}_{d}\left( \sum_{i=1}^{N} w_i \cdot \mathbf{h}_{expert\_i} \right)$$

\paragraph{Classification Module}
The representation $\mathbf{h}_{moe}$ serves as input to the final classification head. This module is implemented as a two-layer MLP that maps the high-dimensional features to the label space. Specifically, the input is projected to a hidden dimension of 20, regularized via Dropout ($p=0.5$) and ReLU activation, before passing through the output layer:$$\mathbf{z} = \text{Linear}_{2}(\text{ReLU}(\text{Dropout}(\text{Linear}_{20}(\mathbf{h}_{moe}))))$$where $\mathbf{z} \in \mathbb{R}^2$ represents the predicted logits for the binary classification task.

\subsection{Model Training and Evaluation Setup}
\paragraph{Pre-training and Initialization}
We employed a fine-tuning strategy based on Methyl-GP to initialize our feature encoders~\cite{xie2025methyl}. For the DNABERT branch, we directly utilized the official pre-trained weights provided by Methyl-GP; for the DNABERT2 branch, we replicated their multi-species fine-tuning protocol. Specifically, we constructed aggregated datasets for 4mC, 5hmC, and 6mA: the former two incorporated training samples from all species, whereas the 6mA dataset excluded \textit{T. thermophile} to mitigate bias arising from its excessive sample size. The DNABERT2 encoder was fine-tuned on these domain-specific datasets. Ultimately, both branches were initialized using the fine-tuned parameters.

\paragraph{Implementation and Training Strategy}
The MEDNA-DFM framework was implemented using PyTorch (v1.13.1) and the Transformers library (v4.18.0)~\cite{paszke2019pytorch, 2020transformers}, with all experiments conducted on NVIDIA A100 80GB GPUs. We employed the AdamW optimizer~\cite{loshchilov2018decoupled} ($\beta_1=0.9, \beta_2=0.999$) to minimize a composite objective function that integrates flooding regularization and adversarial training. To mitigate overfitting, we adopted a flooding-regularized cross-entropy loss. Let $$\mathcal{L}_{CE} = - \sum_{k=0}^{1} y_k \log(\hat{y}_k)$$ denote the standard cross-entropy loss for the binary classification task; the final optimized loss $\mathcal{L}$ is defined as:$$\mathcal{L} = |\mathcal{L}_{CE} - b| + b$$where $b$ is the dataset-specific flood level. To further enhance model robustness, we incorporated adversarial training via the Fast Gradient Method (FGM)~\cite{LIU2023101697}. Specifically, we applied a perturbation $\mathbf{r}{adv}$ to the embedding parameters $\mathbf{w}$ during the forward pass, scaled by the gradient's $\ell_2$-norm with $\epsilon = 1.0$:$$\mathbf{r}_{adv} = \epsilon \cdot \frac{\nabla_{\mathbf{w}} \mathcal{L}(\theta, \mathbf{x}, y)}{\|\nabla_{\mathbf{w}} \mathcal{L}(\theta, \mathbf{x}, y)\|_2}$$Model parameters were updated using accumulated gradients from both the original and perturbed passes. Key hyperparameters—including learning rate, batch size, weight decay, and flood level $b$—were optimized individually for each of the 17 benchmark datasets to maximize the mean AUPRC under 5-fold cross-validation.

\paragraph{Evaluation Metrics}
Defining methylated sites as positive instances, we evaluated model performance using a fixed decision threshold of 0.5. Based on this cutoff, we reported Accuracy (ACC), Sensitivity (Sn), Specificity (Sp), and the Matthews Correlation Coefficient (MCC). Additionally, to assess discriminative power independent of the decision boundary, we calculated the Area Under the Receiver Operating Characteristic Curve (AUC). They provide a quantitative measure of the model's global performance, where higher scores are indicative of better classification quality.

\begin{align*}
    ACC &= \frac{TP + TN}{TP + TN + FP + FN} \\
    Sn &= \frac{TP}{TP + FN} \\
    Sp &= \frac{TN}{TN + FP} \\
    MCC &= \frac{TP \times TN - FP \times FN}{\sqrt{(TP + FP)(TP + FN)(TN + FP)(TN + FN)}}
\end{align*}

\noindent where $TP$ (True Positive) denotes the number of methylated sites correctly identified by the model; $TN$ (True Negative) represents the number of non-methylated sites correctly identified; $FP$ (False Positive) refers to the number of non-methylated sites incorrectly predicted as methylated; and $FN$ (False Negative) indicates the number of methylated sites incorrectly predicted as non-methylated.

\paragraph{Ablation Study Design}
To dissect the contribution of each component within MEDNA-DFM, we conducted a series of ablation studies. Unless otherwise specified, all variants retained the same architecture, hyperparameters, and training procedures as the final model.

\textbf{Optimization of $k$-mer Granularity.} To determine the optimal tokenization resolution, we evaluated four variants by initializing the DNABERT encoder with weights pre-trained on different $k$-mer lengths ($k \in \{3, 4, 5, 6\}$).

\textbf{Necessity of Asymmetric Modulation.} To validate the specific FiLM-based architecture, we compared the final model against three structural variants. The \textbf{Single-Encoder Baseline} excluded the DNABERT2 branch entirely, utilizing only the $\mathbf{h}_{kmer}$ representation for classification. The \textbf{Concatenation Fusion} variant substituted the element-wise affine transformation with a direct concatenation of $\mathbf{h}_{kmer}$ and $\mathbf{h}_{bpe}$. The \textbf{Reverse Modulation} variant inverted the dependency, employing $\mathbf{h}_{kmer}$ to generate the affine parameters that modulate $\mathbf{h}_{bpe}$.

\textbf{Configuration of the MoE Module.} We validated the Mixture-of-Experts strategy by first testing a No-MoE baseline, where the modulated features were directly fed into the classification head. Subsequently, to determine the optimal expert capacity, we evaluated variants with varying numbers of experts ($N \in \{1, 2, 4, 8\}$), analyzing the trade-off between model complexity and predictive performance.

\textbf{Impact of Training Strategies.} We quantified the impact of our specialized training protocols through two control experiments: (i) w/o Adversarial Training: A variant disabling the Fast Gradient Method (FGM) perturbation to assess its contribution to robustness. (ii) w/o Specific Fine-tuning: A variant initializing the Dual-View module with off-the-shelf pre-trained DNABERT weights, bypassing our domain-adaptive fine-tuning process.

\subsection{Feature Representation Visualization}
To visualize the evolution of learned representations, we extracted 768-dimensional feature vectors from the test set at four distinct computational stages of the MEDNA-DFM inference process. These high-dimensional vectors were projected into a two-dimensional embedding space using the Uniform Manifold Approximation and Projection algorithm~\cite{Healy2024}, implemented via the umap-learn library (v0.5.9). The projection was configured with n\_neighbors=100, min\_dist=0.1, and random\_state=42 to ensure stability. The visualization targets the following representations: (1) the [CLS] token output ($h_{kmer}$) from the randomly initialized DNABERT-6mer encoder; (2) the $h_{kmer}$ from the fully fine-tuned encoder; (3) the modulation output from the FiLM module ($h_{mod}$); and (4) the final aggregated output from the Mixture of Experts module ($h_{moe}$).

\subsection{Evaluation for Cross-Species Generalization}
To evaluate the transferability of the learned features, we conducted an exhaustive pairwise evaluation among the 17 benchmark datasets. We constructed a performance matrix where each cell $(i, j)$ denotes the inference performance (AUC) of the predictor trained on dataset $i$ when applied to the test set of dataset $j$. This process involved direct inference using the frozen parameters of the source model. The raw AUC matrix was subsequently row-normalized by dividing each value by the corresponding on-diagonal AUC (source-on-source performance) to derive the relative performance ratios visualized in the study.

\subsection{Contrastive Signal Purification Algorithms}
The design of our CWGA and CAD algorithms is grounded in a dual-source validation: the mechanistic dissection of the model's internal decision-making (detailed in Supplementary Note 1) and the critical recognition that raw attention weights often produce 'artifacts' rather than faithful explanations~\cite{jain2019attention}. To address these limitations, we formulated these algorithms to mitigate statistical noise and align interpretation with the model's actual feature modulation strategies, ensuring the identification of genuine biological drivers.

\paragraph{Contrastive Attention Cohen's d (CAD)}

To identify the motifs most diagnostic for discriminating between positive and negative samples within the Dual-View DNABERT module, we designed and implemented a post-hoc statistical procedure termed CAd (Contrastive Attention Cohen's d). This procedure was developed to quantify the effect size of the attentional differences directed toward specific motifs between the two sample classes.

\textbf{Generation of high-confidence sample sets:} The analysis began with the construction of high-confidence contrastive sample sets. We loaded the pre-trained MEDNA-DFM model to perform predictions on the entire test set. Sequences were then filtered to include only those that were both correctly classified and achieved a Softmax confidence score greater than 0.9 ($\text{confidence\_threshold}$). This process yielded a high-confidence positive sample set ($S_{pos}$) and a high-confidence negative sample set ($S_{neg}$).

\textbf{Construction of attention score distributions:} Next, the DNABERT-6mer and DNABERT2 modules were loaded. We then iterated through all sequences in both the $S_{pos}$ and $S_{neg}$ sets to extract attention scores. Specifically, for each sequence, the attention scores directed from the [CLS] token to every motif token ($m$) were obtained from the final transformer layer and averaged across all attention heads. For any given motif $m$, all corresponding scores were collected to construct an attention score distribution for the positive sample set, $A_{pos} = \{p_1, p_2, ..., p_{n_{pos}}\}$, and for the negative sample set, $A_{neg} = \{n_1, n_2, ..., n_{n_{neg}}\}$.

\textbf{Calculation of CAd and statistical significance:} Finally, the CAd and its statistical significance were calculated. We set a minimum occurrence threshold ($\text{MIN\_SAMPLES} = 3$), and all motifs $m$ that met this threshold in both distributions (i.e., $n_{pos} \ge 3$ and $n_{neg} \ge 3$) were processed.For each qualifying motif, its Contrastive Attention Cohen's d was calculated as the standardized effect size, representing the difference between the two distribution means ($\bar{A}_{pos} - \bar{A}_{neg}$) divided by their pooled standard deviation $s_p$:$$CAd_m = \frac{\bar{A}_{pos} - \bar{A}_{neg}}{s_p}$$where $\bar{A}_{pos}$ and $\bar{A}_{neg}$ are the means of the $A_{pos}$ and $A_{neg}$ distributions, respectively. The pooled standard deviation $s_p$ was calculated as follows:$$s_p = \sqrt{\frac{(n_{pos} - 1)s_{pos}^2 + (n_{neg} - 1)s_{neg}^2}{n_{pos} + n_{neg} - 2}}$$In this formula, $n_{pos}$ and $n_{neg}$ are the sample sizes of $A_{pos}$ and $A_{neg}$ (i.e., the number of occurrences of motif $m$), and $s_{pos}^2$ and $s_{neg}^2$ are the sample variances of these two distributions, respectively. Concurrently, to assess the statistical significance of the observed CAd values, a Welch's t-test was performed on the two independent distributions, $A_{pos}$ and $A_{neg}$. This test was selected as it is statistically more robust and does not assume equal variances. The final results were sorted in descending order by their CAd values to highlight the most discriminative diagnostic motifs.

\paragraph{Contrastive Weighted Gradient Attribution (CWGA)}

To systematically elucidate the sequence patterns determinative of model decisions in a DNA sequence binary classification task, we designed and implemented the Contrastive Weighted Gradient Attribution (CWGA) method. This approach leverages gradient attribution techniques, is grounded in high-confidence samples, and is weighted by the effect size of feature dimensions. It quantifies the contribution of sequence units at different granularities (BPE tokens and 6mer tokens) to the model's classification outcomes.

\textbf{High-confidence contrastive sample selection:} We utilized the trained MEDNA-DFM model to perform inference on the test set. Samples that were correctly classified with a Softmax confidence score exceeding a predefined threshold were filtered. The top $N$ (NUM\_SAMPLES\_TO\_ANALYZE = 100) samples were then selected to construct the high-confidence positive set $S_{pos}$ and the high-confidence negative set $S_{neg}$, respectively.

\textbf{Discriminative feature dimension selection:} For each feature dimension $d$ output by the FiLM module, we calculated the mean difference in activation values between the positive and negative samples
\[
\Delta_d = \bar{O}_{pos}^d - \bar{O}_{neg}^d
\]
We then employed a two-sided t-test to assess the statistical significance (p-value) of this difference and proceeded to compute the standardized effect size, Cohen's $d$. All dimensions were ranked according to the absolute value of Cohen's $d$, $|d_d|$, and the top $K$ (TOP\_N\_DIMS\_FOR\_AGGREGATION = 40) most discriminative dimensions were selected to form the set $D_{top}$. The resulting $d_d$ value served as the weight for the subsequent attribution calculations.

\textbf{Attribution calculation:} For each selected dimension $d \in D_{top}$, we employed distinct integrated gradient methods to compute the contribution of the two types of sequence units. For BPE tokens, we applied the Layer Integrated Gradients (LIG) method. With the embedding layer as the attribution target, we calculated the attribution scores $A_{pos}^d(t_b)$ and $A_{neg}^d(t_b)$ for each BPE token $t_b$ with respect to the output value of dimension $d$. The difference was taken as the token's discriminative contribution for that dimension:$$\delta_d(t_b) = A_{pos}^d(t_b) - A_{neg}^d(t_b)$$ For 6mer tokens, we applied the standard Integrated Gradients (IG) method. With the embedding vector as the attribution target, we calculated the attribution scores $A_{pos}^d(t_k)$ and $A_{neg}^d(t_k)$ for each 6mer unit $t_k$ with respect to the output value of dimension $d$. The difference was taken as the k-mer's discriminative contribution for that dimension:$$\delta_d(t_k) = A_{pos}^d(t_k) - A_{neg}^d(t_k)$$

\textbf{Weighted aggregation and normalization:} The token contribution $\delta_d(t)$ from each dimension was weighted by that dimension's Cohen's $d$ value, $d_d$. These weighted contributions were then summed over all dimensions in $D_{top}$ to obtain the total contribution $C(t)$ for a given token $t$:$$C(t) = \sum_{d \in D_{top}}(d_d \times \delta_d(t))$$Finally, this total contribution was normalized by the maximum absolute contribution across all tokens:$$\hat{C}(t) = \frac{C(t)}{\max_{t'} |C(t')|}$$The final results were sorted in descending order to highlight the most discriminative diagnostic motifs.

\section{Data availability}
The datasets in this study are available from the corresponding author on reasonable request, and will be made fully public upon the formal publication of this manuscript.

\section{Code availability}
The complete repository will be made publicly available upon the formal acceptance and publication of this article. In the interim, researchers may contact the corresponding author for access to specific algorithmic implementations on a reasonable request basis.

\bibliography{references}

\end{document}